\documentclass[conference]{IEEEtran}
\IEEEoverridecommandlockouts
\usepackage{amsmath,amssymb,amsfonts}
\usepackage{graphicx}
\usepackage{textcomp}
\usepackage{xcolor}

\usepackage[utf8]{inputenc}
\usepackage{amssymb,amsmath,array}
\usepackage{xcolor}
\usepackage{adjustbox}
\usepackage{multirow}
\usepackage{layouts}
\usepackage{booktabs}
\usepackage{biblatex}
\usepackage{makecell}
\usepackage{hyperref}
\usepackage{tikz}
\usepackage{float}
\usepackage{blindtext}
\usepackage{caption} 

\usepackage{algorithm}
\usepackage{algpseudocode}

\bibliography{bib}

\def\BibTeX{{\rm B\kern-.05em{\sc i\kern-.025em b}\kern-.08em
    T\kern-.1667em\lower.7ex\hbox{E}\kern-.125emX}}
\begin{document}

\title{Combining pre-trained Vision Transformers and CIDER for Out Of Domain Detection}



\author{
    \IEEEauthorblockN{Grégor Jouet\IEEEauthorrefmark{1}\IEEEauthorrefmark{2}\IEEEauthorrefmark{4}, Clément Duhart\IEEEauthorrefmark{1}}
    \IEEEauthorblockA{
        \IEEEauthorrefmark{1}Pôle Universitaire Léonard de Vinci, Research Center
        \\La Défense, France
        \\ \{gregor.jouet, clement.duhart\}@devinci.fr
        \\    
    }
    \and
    
    \IEEEauthorblockN{Julio Laborde\IEEEauthorrefmark{2}}
    \IEEEauthorblockA{
         \IEEEauthorrefmark{2}reciTAL
         \\Paris, France
         \\ \{julio, gregor\}@recital.ai
    }
    \and
    \IEEEauthorblockN{Francis Rousseaux\IEEEauthorrefmark{3}}
    \IEEEauthorblockA{
        \IEEEauthorrefmark{3}URCA CReSTIC, Moulin de la Housse
        \\Reims, France
        \\francis.rousseaux@univ-reims.fr
    }
    \and
    \IEEEauthorblockN{Cyril de Runz\IEEEauthorrefmark{4}}
    \IEEEauthorblockA{
        \IEEEauthorrefmark{4}University of Tours, LIFAT, BdTLN
        \\ Tours, France
        \\ cyril.derunz@univ-tours.fr, gregor.jouet@etu.univ-tours.fr
    }
}

\maketitle

\begin{abstract}
Out-of-domain (OOD) detection is a crucial component in industrial applications as it helps identify when a model encounters inputs that are outside the training distribution. Most industrial pipelines rely on pre-trained models for downstream tasks such as CNN or Vision Transformers. This paper investigates the performance of those models on the task of out-of-domain detection. Our experiments demonstrate that pre-trained transformers models achieve higher detection performance out of the box. Furthermore, we show that pre-trained ViT and CNNs can be combined with refinement methods such as CIDER to improve their OOD detection performance even more. Our results suggest that transformers are a promising approach for OOD detection and set a stronger baseline for this task in many contexts.
\end{abstract}

\begin{IEEEkeywords}
out-of-domain, vision transformer, transformer, cnn
\end{IEEEkeywords}

\section{Introduction}

Out-of-domain (OOD) detection, the capability of a machine learning model to correctly identify samples that are not part of the data distribution it was trained on, has become an increasingly important research topic~\cite{hendrycksScalingOutofDistributionDetection2022} in the field of computer vision, especially for image classification tasks. With the recent advances in deep learning techniques~\cite{vaswaniAttentionAllYou2017} and the emergence of powerful transformer-based vision models such as ViT~\cite{dosovitskiyImageWorth16x162021}, the need for OOD detection has become even more pronounced. This need for a system to detect OOD samples goes hand in hand with the need for trust systems for recent deep learning application and is shown by the emergence of fields like explainable AI~\cite{islamExplainableArtificialIntelligence2021} or uncertainty estimation~\cite{greenMACEstReliableTrustworthy2021, abdarReviewUncertaintyQuantification2021}.

In the image classification context, OOD detection is essential to prevent misclassification or model failure in real-world scenarios where images may differ from those in the training dataset.~\cite{nguyenDeepNeuralNetworks2015b} For example, if a self-driving car is trained to recognize stop signs only under specific lighting conditions, it may fail to detect a stop sign when presented with different lighting conditions. This could have catastrophic consequences. Moreover, detecting OOD images accurately is also important for ensuring the robustness and reliability of AI systems~\cite{royDoesYourDermatology2021, bleiIdentifyingOutofDistributionSamples2022, sehwagSSDUnifiedFramework2021}. 

Given the critical importance of OOD detection, there is a growing need to seek better out-of-the-box performance for out-of-domain detection. This refers to the ability of a model to detect OOD images with high accuracy without any specific fine-tuning to improve OOD detection. With the development of newer and more complex models, achieving better OOTB performance has become an even more pressing concern~\cite{yangOpenOODBenchmarkingGeneralized2022}. However, there it currently no empirical exploratory work on the use of pre-trained Vision Transformer models for Out of Domain detection nor research on their improvement using OOD refinement methods.

In light of these considerations, this paper aims to explore the current state of OOD detection in the context of image classification, with a focus on transformer-based models. The paper will examine the various techniques and approaches used to improve OOD detection currently used with CNN-based, as well as the challenges and limitations of current approaches. We will especially use the CIDER~\cite{mingCIDERExploitingHyperspherical2022} method which uses hypershperical embeddings to separate in from out of domain samples. By highlighting the importance of OOD detection and its relevance to the wider AI landscape, this paper aims to contribute to ongoing efforts to develop more robust and reliable machine learning models for real-world applications.

Convolutional neural networks (CNNs) have long been the standard model family for image processing tasks, with state-of-the-art performances on a wide range of benchmarks and datasets. However, with the advent of transformer-based models such as the popular BERT~\cite{devlinBERTPretrainingDeep2019} and GPT~\cite{brownLanguageModelsAre2020}, there has been a growing interest in exploring the potential of these models for image classification tasks. While transformers have shown impressive performance in natural language processing tasks, their application to image processing is still a relatively new area of research.~\cite{baoBEiTBERTPreTraining2021}

One of the key differences between CNNs and transformers is their underlying architecture. CNNs are designed to learn hierarchical representations of images by applying a series of convolutional and pooling operations, while transformers rely on self-attention mechanisms to process sequences of data. While both models are capable of capturing complex patterns and relationships within images, their approaches differ significantly. 

Another important difference between transformers and CNNs is their computational efficiency. Pre-trained CNNs require less memory than vision transformer models, which rely on attention mechanisms that require more memory. For this reason, ViT models are heavier and more difficult to train but offer better features than CNN and are the current state of the are in a lot of computer vision domains.~\cite{baoBEiTBERTPreTraining2021} 

In this paper, we aim to provide a comprehensive analysis of the differences between transformers and CNNs when using them for OOD detection with an underlying image classification task. We will compare and contrast the two types of models in terms of their architecture and performance on a range of OOD detection methods. By examining the strengths and limitations of each model, we hope to shed light on the potential of transformer-based models for image OOD detection in image processing pipelines and pave the way for future research in this area. 

To summarize, our contributions are as follows:
\begin{enumerate}
  \item We empirically demonstrate that fine-tuning a Vision Transformer model yields superior out-of-domain detection performance compared to a Convolutional Neural Network on a variety of OOD methods.
  \item We use the recent CIDER method for OOD detection performance improvement on pre-trained convolutional models.
  \item Further, we apply the same CIDER method to a transformer-based architecture, resulting in improved OOD performance.
\end{enumerate}
In the rest of the paper, we first present a comprehensive review of the current state of the art on OOD detection methods and OOD detection performance improvements of convolutional models. We then formally introduce our pipeline and experiments in Section~\ref{sec:ood_method} and then present our results in Section~\ref{sec:ood_results} and conclude.

\section{Related Work}

In this section, we will explore various Out of Domain detection methods that are commonly used in a variety of contexts. We will then focus on methods to enhance the performance of Out of Domain detection when using Convolutional Neural Networks (CNN). 
\subsection{OOD evaluation methods}
\label{subsec:ood:usual_methods}
\textbf{MaxSoftmax}~\cite{hendrycksBaselineDetectingMisclassified2018a} - The MaxSoftmax method was the first approach used to identify out of domain samples. This technique involves using the maximum softmax value of a classifier for an example and marking it as OOD if the value is below a certain threshold. A similar approach, known as the MaxLogit approach, uses the raw logit value instead of the softmax. These methods are still utilized in certain scenarios, mainly under the assumption that the classifier is well calibrated. They have been employed for a considerable amount of time and remain an essential baseline method.

\textbf{Mahalanobis}~\cite{manahalobis2018} - Out Of Domain detection can also be achieved through projecting a sample into a high dimensional space. This approach relies on the assumption that samples belonging to the same class should be close to each other and far from all other prototypes. The Mahalanobis method is a popular example of this technique, which calculates the Mahalanobis distance between the sample distribution of each class and a new example in order to classify it as either in or out of domain. The high dimensional space may be from a multi layer perceptron projector or the output of a pre-trained model such as a vision or language model.

\textbf{Energy Based}~\cite{liuEnergybasedOutofdistributionDetection2021} - Energy Based Methods utilize the Energy Based Model (EBM) paradigm to assess the out of domain nature of a sample. The EBM maps the model input to a scalar, known as the energy, which is determined when constructing the model. Generally, the energy for known examples is higher than that of unseen examples. A threshold $\tau$ is chosen, and samples with an energy lower than this value are deemed Out of Domain. The threshold is typically chosen using in domain and out of domain data so that the OOD classifier has a high fraction of correct classification.

\textbf{ODIN}~\cite{liangEnhancingReliabilityOutofdistribution2020a} - \emph{Out-of-Distribution detector for Neural networks} (ODIN) is an OOD detection method that relies on temperature scaling and a custom data pre-processor to add a small perturbation to input images. ODIN then calculates the softmax classification value for both the original image and the perturbed image and classifies the input if the difference between max softmax values is above a certain threshold. The underlying idea, which originates from adversarial examples, is to make the network wrongly classify examples. This is because the network reacts more strongly when noise is added to an in-domain example than to an out of domain one. 


\textbf{KLMatching}~\cite{hendrycksScalingOutofDistributionDetection2022} - Hendrycks et al. first fit a typical posterior distribution $d_y$ computed for each class: $d_y = \mathbb{E}_{x \sim \mathcal{X}_{val}}[p(y \vert x)]$ and at inference, computer the KL between the given sample and the typical posterior: $D_{KL}[p(y \vert x) \Vert d_y]$ The KL is used as an outlier score and if greater than a predefined threshold, the sample is considered to be OOD. This method has the advantage of not requiring the labels of the samples as we are only using the posterior value. The KL Matching method also works in multi-class contexts.

\textbf{OpenMax}~\cite{bendaleOpenSetDeep2015} - The OpenMax model introduces a new layer creating a centroid $\mu_j$ in logit space for each class using Extreme Value Theory (EVT) and yields an outlier value by computing the softmax value of a revised activation vector chained by the Weibull CDF probability. A more recent application of EVT is the Extreme Value Machine~\cite{ruddExtremeValueMachine2018}, which constructs a decision boundary using EVT.

\subsection{Convolutional OOD improvement}
\label{subsec:ood:improve}
Aside from traditional out-of-domain (OOD) detection methods, there are a variety of other methods present in the literature to improve the performance of OOD detection. These methods often rely on the incorporation of a regularization loss component or a deep representation of the data in order to find a projection space with desirable properties for out of domain detection algorithms.

\textbf{VOS}~\cite{duVOSLearningWhat2022} - \emph{Virtual Outlier Synthesis} is a method that improves OOD (Out-of-Domain) performances by introducing out of domain samples (outliers) into the representation space. Gaussian representations of in-domain data for each class are combined with these outliers to compute a regularization loss component, $\mathcal{L}_{uncertainty}$, which is derived from the free energy. This component has been shown to improve OOD detection. Ultimately, the method optimizes for the separability of OOD and ID (In-Domain) samples in the representation space. 

\textbf{CIDER}~\cite{mingCIDERExploitingHyperspherical2022} - \emph{CIDER} is a recent method in the literature that uses hyperspherical embeddings on which it project the deep representation of samples. These samples are then pushed away from each other depending on their classes while the samples from a same class are regrouped together. This is accomplished with two different regularization losses for the regrouping and spreading of samples. The spherical nature of the feature space helps to this end by making the classes more separable. CIDER thus relies on the geometry of the feature space as well as two regularization loss components. The network is separated in two parts: a backbone to extract features from the images followed by a projection head, projecting those features on the hypersphere. Once the CIDER training is complete, the head network is replaced with a linear probe to evaluate the model in-domain performance. 


\subsection{Discussion}

This state-of-the-art showcases traditional methods for Out Of Domain Detection as well two methods used to improve the OOD performances of the methods in section~\ref{subsec:ood:usual_methods}. We note however that methods presented in section~\ref{subsec:ood:improve} only cover the case of CNN backbone models. The literature, to the best of our knowledge, does not have such methods for classification models with a Vision Transformer or any Transformer-based backbone. 

This lack of literature could be explained by the natural ability of pre-trained vision transformers at OOD detection or because the concepts developed for convolutional neural networks do not transfer well to transformer-based models. Indeed, ViT are architecturally different from CNNs and could maybe capture more relevant features for OOD detection tasks. It is also possible ViT pre-training procedure helps them capture better features. The methods employed on CNN models could have a dependency on the convolutional nature of these models and not be applicable to ViT. We explore these two hypotheses in this paper and detail what questions we ask regarding this state of the art in Section~\ref{ood:research_questions}.

\section{Method}  
\label{sec:ood_method}

% ...

In this section, we first formalize the problem setup and introduce our research questions. We then present the different experimental conditions for our work and the different pipelines we have put in place. Finally, we go into detail about the experimental conditions and implementation of our work.

\subsection{Problem formalization}

Our study is on deep learning models specific to image classification. Formally, an image classification model is a mapping from $[0, 1]^{wxh}$ to $[[1, C]]$, where $w, h$ are the width and height of the input image and $C$ is the number of target classes. 
In our case, this mapping consists of a backbone neural network that extracts meaningful features from the input image and a classification head that outputs the logits of each class. The maximum logit index is then chosen as the predicted class for the example. The overall model can be expressed as $\mathcal{M}_c = x \rightarrow classifier(backbone(x))$, where $x$ is the input image. The output of the backbone network is referred to as the feature space with a chosen dimensionality of $F_d \in \mathbb{N}$. 
In addition to $\mathcal{M}_c$, we also use the CIDER method on our $backbone$. When used, we change the classifier to a projection head, which yields a projection of a chosen dimension $F_{h}$. The projection head is a mapping $\mathbb{R}^{F_{d}} \rightarrow  \mathbb{R}^{F_{h}}$. This projection head is used to train the model using the CIDER method.
The $backbone$ network can either be a CNN model or ViT model. The different models used and their size are detailed in Tables~\ref{ood:cnn_variant_table} and~\ref{ood:vit_variant_table}. 
The evaluation uses the methods described in Section~\ref{subsec:ood:usual_methods} applied to the classifier model $\mathcal{M}_c$

\subsection{Research Questions} 
\label{ood:research_questions}

This work aims to investigate the potential of pre-trained vision transformer models for out-of-domain use cases. To this end, we pose the following research questions: 
\begin{enumerate}
  \item What is the baseline performance of pre-trained ViT models on classical OOD datasets with methods described in section~\ref{subsec:ood:usual_methods} compared to pre-trained convolutional neural network (CNN) models?
  \label{ood_research_question:1}
  \item Is the CIDER method applicable to pre-trained convolutional models and what result does it yield?
  \label{ood_research_question:2}
  \item Are the methods from CIDER transferable to pre-trained transformer-based models and do they improve OOD performance? 
  \label{ood_research_question:3}
\end{enumerate}



\begin{figure}[ht]
    \includegraphics[width=0.5\textwidth]{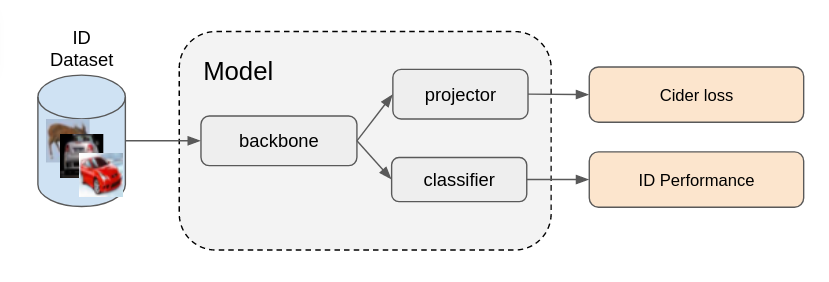}
    \caption{Representation of the pipeline used in our experiments with the CIDER method. The \textit{backbone} model can be an untrained CNN, pre-trained CNN or a pre-trained ViT model. We first train the model using the CIDER method, then train a linear classification probe and evaluate the model using different OOD datasets.}
    \label{ood:fig:pipeline}
\end{figure}

\begin{algorithm}
\caption{The CIDER fine-tuning pipeline used in our experiments in the \textit{CIDER condition}. The backbone can be any specific model of the CNN or ViT architecture class.}\label{alg:general_cider_training_pipeline}
\begin{algorithmic}
    \Require \textit{backbone}: $[0,1]^{w \textsubscript{x} h} \rightarrow \mathbb{R}^{F_{d}}$ 
    \Require \textit{head}: $\mathbb{R}^{F_{d}} \rightarrow \mathbb{R}^{F_{h}}$
    \Require $\mathcal{D}_{train}$ train dataset

    \ForAll{$(x_i, y_i) \in \mathcal{D}_{train}$}
        \State $d_c = normalize(backbone(x_i))$
        \State $d_p = head(d_c)$
        \State Compute $\mathcal{L}_{dis}$ and $\mathcal{L}_{comp}$
        \State Do a learning step.
    \EndFor 
    \State\algorithmicreturn~the trained \textit{head}
\end{algorithmic}
\end{algorithm}

To answer our \hyperref[ood_research_question:1]{first research question}, we develop an OOD testing pipeline composed of a convolutional or vision transformer backbone and a linear probe. We finetune the model on the \textit{Cifar10} in-domain dataset and evaluate its OOD performance using different datasets and methods. To answer questions~\ref{ood_research_question:2} and~\ref{ood_research_question:3}, we first fine-tune models using the CIDER method, then freeze the backbone model and test its performance with a linear probe, finally evaluating it on the same benchmark used for question~\ref{ood_research_question:1}. 

\begin{figure}[ht]
    \includegraphics[width=0.5\textwidth]{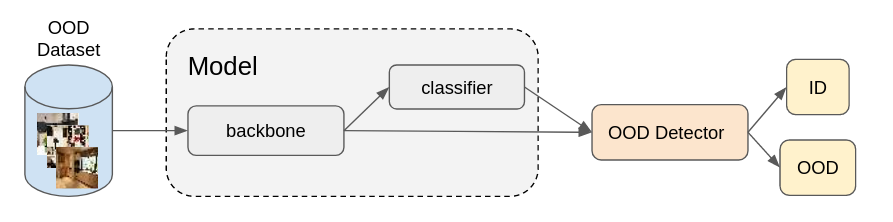}
    \caption{Model evaluation on OOD detection task using the trained classifier except for the \textit{Manalahobis} method which needs the deep representation of the samples. The OOD detector block represents the different methods described in Section~\ref{sec:ood_method}} 
    \label{ood:fig:pipeline_ood}
\end{figure}

\subsection{Pre-trained ViT vs CNN}

Our main goal is to fairly assess the OOD performances of pre-trained ViT models and CNN. The CNN models rely on convolution kernels to automatically extract features and build a hierarchical representation of features across layers. Vision Transformers  use attention layers to create a mapping between different elements of an image and create a good representation. The architectures of the models are different and they create different representations of the inputs. For this reason, the comparison between the models is difficult to realize in a fair manner. We explore different settings to ensure a reasonable comparison between the models: We choose different model sizes for each model family to compare the number of parameters and we also report the in-domain performances the models. 
In this section, we adopt a very simple testing pipeline by first fine-tuning the models on an in-domain dataset and then test different methods to identify OOD examples. 
The OOD detection pipeline used in all our experiments after the finetuning phase is detailed in Figure~\ref{ood:fig:pipeline_ood}.
We detail the different model size in Section~\ref{subsec:experimental_conditions} and the in domain performance in Section~\ref{sec:ood_results}.


\subsection{CIDER applied to Pre-trained CNN}
As we have seen the CIDER method requires to build class prototype before calculating the outlier score during inference. The CIDER paper details how these prototypes are built using a combination of losses: the \textit{dispersion} ($\mathcal{L}_{dis}$) and \textit{compactness} ($\mathcal{L}_{comp}$) losses. These losses are applied throughout the training of the model and, in fact, replace the traditional Cross Entropy loss. CIDER effectively trains model from scratch using the aforementioned losses. 

Our \hyperref[ood_research_question:2]{second research question} focuses on the application of the CIDER method on pre-trained CNNs, we want to know if the CIDER method improved the OOD performance of a pre-trained CNN model and what are the performance implications on the model. To answer both questions, we adopt a two stages pipeline: We first remove the classification head of the backbone to obtain deep representations of images of dimension $F_d \in \mathbb{N}$. As required by CIDER, we normalize these features and obtain $d_{c} = normalize(backbone(x))$. We then pass these normalized features to a projection head $d_{p} = head(d_{c})$ and finally use the CIDER losses $\mathcal{L}_{dis}$ and $\mathcal{L}_{comp}$ to train the model. 
We use Algorithm~\ref{alg:general_cider_training_pipeline} for the first stage of this pipeline and the Algorithm~\ref{alg:general_cider_evaluation} for evaluation using a linear probe. We provide an illustration of this pipeline in figure~\ref{ood:fig:pipeline} 

\begin{algorithm}
    \caption{Cider evaluation pipeline. We use a linear probe to evaluate the model following~\cite{mingCIDERExploitingHyperspherical2022}}\label{alg:general_cider_evaluation}
    \begin{algorithmic}
        \Require \textit{backbone}: $[0,1]^{w \textsubscript{x} h} \rightarrow \mathbb{R}^{F_{d}}$ 
        \Require \textit{probe}: $\mathbb{R}^{F_{d}} \rightarrow [0,1]^{C}$ a linear layer
        \Require $\mathcal{D}_{train}$ train dataset
        \Require $\mathcal{D}_{test}$ test dataset
    
        \State Freeze \textit{backbone} 
        \State Initialize \textit{probe}
        \State Train the model $x \rightarrow probe(normalize(backbone(x)))$ on $\mathcal{D}_{train}$
        \State Evaluate the model on $\mathcal{D}_{test}$

        \State\algorithmicreturn~the evaluation metrics
    \end{algorithmic}

\end{algorithm}

\subsection{CIDER applied to Pre-trained Vision Transformers}
Until now we have used CIDER only on CNN models to improve their OOD performance. The \hyperref[ood_research_question:3]{third research question} asks if the CIDER method can be applied to Vision Transformer models. More specifically, we are interested in if CIDER improves the OOD performances of pre-trained ViT. 
To answer this question, we use the same pipeline and Algorithm~\ref{alg:general_cider_training_pipeline} with a pre-trained ViT backbone. And Algorithm~\ref{alg:general_cider_evaluation} for evaluation. This way we can also have comparable results across architectures, with limitations detailed in Section~\ref{sec:ood_results}. We adjust the head dimensions depending on the backbone's $F_d$ output dimension. For instance, if the ViT model has an output feature dimension of $F_d = 768$, then the heads will have a dimension of 768. 

The ViT model provides a good deep representation for the CIDER module to use. We adjust the CIDER hyperparameters as detailed in Section~\ref{subsec:experimental_conditions} but keep the same pipeline and architecture. We find that the CIDER module can be applied to pre-trained ViT models without any major modifications. We will detail in Section~\ref{sec:ood_results} the results of this implementation. Although simple, to the best of our knowledge, we are the first to apply this sort of method to pre-trained ViT models.

\subsection{Experimental conditions and details}
\label{subsec:experimental_conditions}

\begin{figure}[h!]
    \centering
    \includegraphics[width=0.5\textwidth]{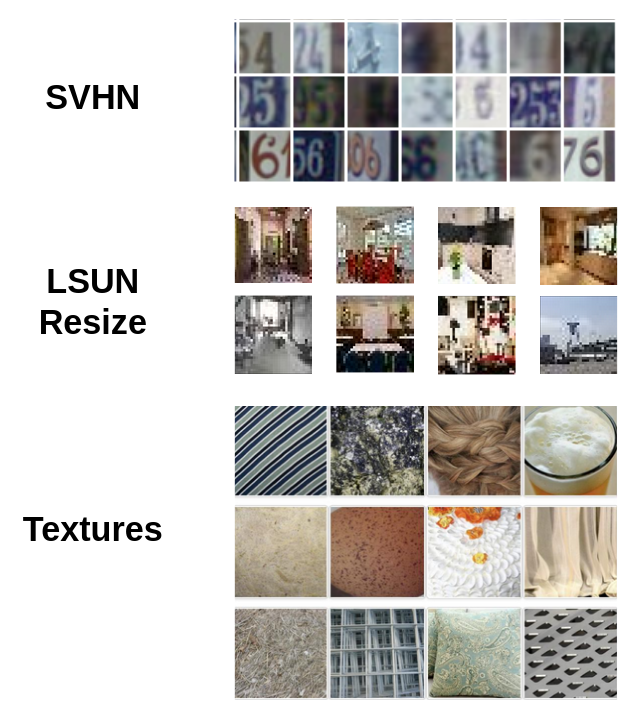}
    \caption{Samples of some of the OOD datasets used in evaluation.}
    \label{ood:fig:examples}
\end{figure}

In our experiments, we use a pre-trained \textit{resnet18} as convolutional backbone and a pre-trained \textit{vit\_b\_32} as a vision transformer backbone.\footnote{Both pretrained backbones are obtained from the torchvision python package.} The OOD datasets used are \textit{SVHN}, \textit{Cifar100} and cropped and resized versions of \textit{LSUN} and \textit{TinyImageNet}. We provide some samples of the datasets used in evaluation in Figure~\ref{ood:fig:examples}

We use two different pipelines to carry out our experiments. In one experimental condition, we load pre-trained models from their checkpoints and finetune them on the in-domain dataset. Then, we test their OOD detection performance on a variety of datasets and methods. This simple approach serves as a baseline for comparisons and, to the best of our knowledge, is the first to compare CNN and ViT in this way. 
The second pipeline, the CIDER pipeline, follows the same architecture as CIDER: we use the pre-trained model feature extractor followed by a high dimension projection head to apply the CIDER losses. The key difference here is that we use a pre-trained model instead of training the model from scratch.

\begin{table}[h!]
    
    \centering
    \begin{tabular}{c|c} 
        \textbf{Model variant (CNN)} & \textbf{Parameters count} \\
        \hline
        $Resnet_{18}$ & $1.1~10^{7}$  \\ 
        $Resnet_{34}$ & $2.1~10^{7}$ \\ 
        $Resnet_{50}$ & $2.5~10^{7}$ \\ 
        $Resnet_{101}$ & $4.4~10^{7}$ \\ 
        $Resnet_{152}$ & $6.0~10^{7}$ \\
    \end{tabular}
    \caption{Number of parameters for each model variant used with a convolutional architecture.}
    \label{ood:cnn_variant_table}
\end{table}

\begin{table}[h!]
    
    \centering
    \begin{tabular}{c|c} 
        \textbf{Model variant (ViT)} & \textbf{Parameters count} \\
        $ViT_{L32}$ & $3.0~10^{8}$ \\
        $ViT_{L16}$ & $3.0~10^{8}$ \\
        $ViT_{B32}$ & $8.7~10^{7}$  \\
        $ViT_{B16}$ & $8.5~10^{7}$ \\
    \end{tabular}
    
    \caption{Number of parameters for each model variant used with a transformer-based architecture.}
    \label{ood:vit_variant_table}
\end{table}


\section{Results} 
\label{sec:ood_results}

\begin{table*}[!h]
    \centering
    \begin{adjustbox}{max width=\textwidth}
        \begin{tabular}{@{}lc*{10}cc@{}}
            \toprule
            \multirow{2}[3]{*}{\bfseries Dataset/Method} & 
            \multicolumn{2}{c}{\makecell{\bfseries $Resnet_{18}$ }} & 
            \multicolumn{2}{c}{\makecell{\bfseries $Resnet_{34}$ }} & 
            \multicolumn{2}{c}{\makecell{\bfseries $ViT_{L16}$ }} & 
            \multicolumn{2}{c}{\makecell{\bfseries $ViT_{L32}$ }} \\ 
            \cmidrule(lr){2-3}\cmidrule(lr){4-5}\cmidrule(lr){6-7}\cmidrule(lr){8-9}
            & AUROC$\uparrow$ & ACC95TPR$\uparrow$ & AUROC$\uparrow$ & ACC95TPR$\uparrow$ & AUROC$\uparrow$ & ACC95TPR$\uparrow$ & AUROC$\uparrow$ & ACC95TPR$\uparrow$ \\
            \midrule    
            Mahalanobis  & 2.51           & 83.59          & 4.71           & 83.59          & 96.94          & 89.71          & \textbf{85.57} & \textbf{90.96}    \\
            MaxLogit     & 82.44          & 91.08          & 76.01          & 89.56          & 97.19          & 90.67          & 85.36          & 88.89    \\
            MaxSoftmax   & 84.81          & 90.71          & 81.48          & 89.56          & 93.99          & 86.92          & 82.80          & 87.99    \\
            ODIN         & 83.19          & 87.99          & 78.83          & 87.99          & 91.96          & 36.06          & 71.99          & 87.99    \\
            OpenMax      & \textbf{95.35} & \textbf{93.45} & \textbf{93.26} & \textbf{91.10} & 95.67          & 89.36          & 85.19          & 89.22    \\
            EnergyBased  & 82.05          & 91.07          & 75.41          & 89.54          & \textbf{97.34} & \textbf{90.86} & 85.39          & 88.89    \\
            KLMatching   & 80.82          & 86.84          & 77.11          & 86.72          & 86.37          & 64.73          & 75.50          & 90.17    \\
            \bottomrule
        \end{tabular}
    \end{adjustbox}
    \caption{Comparison of different OOD detection baselines with different models. The CIDER method is not applied and the results of different variants of the two model families are reported for the SVHN dataset.}
    \label{table:results_ood_first_question}
\end{table*}

\begin{table*}[!h]
    \centering
    \begin{adjustbox}{max width=\textwidth}
        \begin{tabular}{@{}lc*{10}cc@{}}
            \toprule
            \multirow{2}[3]{*}{\bfseries Dataset/Method} & 
            \multicolumn{2}{c}{\makecell{$Resnet_{18}$ scratch }} &  
            \multicolumn{2}{c}{\makecell{$Resnet_{18}$ pre-trained}} & 
            \multicolumn{2}{c}{\makecell{$Resnet_{18}$ + CIDER}} \\ 
            \cmidrule(lr){2-3}\cmidrule(lr){4-5}\cmidrule(lr){6-7}
            & AUROC$\uparrow$ & ACC95TPR$\uparrow$ & AUROC$\uparrow$ & ACC95TPR$\uparrow$ & AUROC$\uparrow$ & ACC95TPR$\uparrow$ \\
            \midrule    
            Mahalanobis  & 5.38           & 83.59          & 2.51           & 83.59          & 54.71          & 83.99    \\
            MaxLogit     & \textbf{80.83} & 90.19          & 82.44          & 91.08          & 75.44          & 88.21    \\
            MaxSoftmax   & 77.43          & 88.98          & 84.81          & 90.71          & 74.47          & \textbf{87.91}    \\
            ODIN         & 80.01          & 87.99          & 83.19          & 87.99          & 73.51          & 85.08    \\
            OpenMax      & 85.66          & 88.06          & \textbf{95.35} & \textbf{93.45} & \textbf{78.29} & 87.70    \\
            EnergyBased  & 80.75          & \textbf{90.23} & 82.05          & 91.07          & 74.76          & 88.25    \\
            KLMatching   & 73.28          & 85.27          & 80.82          & 86.84          & 73.16          & 87.64    \\
            \bottomrule
        \end{tabular}
    \end{adjustbox}
    \caption{OOD detection measurements fot the $Resnet_{18}$ model variant in different conditions: trained from a randomly initialized model (the \textit{scratch} condition), finetuned from a pre-trained model, the \textit{pre-trained} condition and then with then CIDER method applied, the \textit{CIDER} condition. The results are reported for the SVHN dataset.}
    \label{table:results_ood_question2}
\end{table*}

\begin{table*}[!h]
    \centering
    \begin{adjustbox}{max width=\textwidth}
        \begin{tabular}{@{}lc*{10}cc@{}}
            \toprule
            \multirow{2}[3]{*}{\bfseries Dataset/Method} & 
            \multicolumn{2}{c}{\makecell{$Resnet_{18}$ pre-trained }} &  
            \multicolumn{2}{c}{\makecell{$Resnet_{18}$ + CIDER }} & 
            \multicolumn{2}{c}{\makecell{$ViT_{L16}$ }} & 
            \multicolumn{2}{c}{\makecell{$ViT_{L16}$ + CIDER}} \\ 
            \cmidrule(lr){2-3}\cmidrule(lr){4-5}\cmidrule(lr){6-7}\cmidrule(lr){8-9}
            & AUROC$\uparrow$ & ACC95TPR$\uparrow$ & AUROC$\uparrow$ & ACC95TPR$\uparrow$ & AUROC$\uparrow$ & ACC95TPR$\uparrow$ & AUROC$\uparrow$ & ACC95TPR$\uparrow$ \\
            \midrule    
            Mahalanobis  & 2.51           & 83.59          & 54.71          & 83.99          & 96.94          & 89.71          & 97.48          & 93.90    \\
            MaxLogit     & 82.44          & 91.08          & 75.44          & \textbf{88.21} & 97.19          & 90.67          & \textbf{99.34}          & \textbf{95.33}    \\
            MaxSoftmax   & 84.81          & 90.71          & 74.47          & 87.91          & 93.99          & 86.92          & 99.14          & 95.26    \\
            ODIN         & 83.19          & 87.99          & 73.51          & 85.08          & 91.96          & 36.06          & 93.21          & 92.18    \\
            OpenMax      & \textbf{95.35} & \textbf{93.45} & \textbf{78.29} & 87.70          & 95.67          & 89.36          & 97.13          & 95.21    \\
            EnergyBased  & 82.05          & 91.07          & 74.76          & 88.25          & \textbf{97.34} & \textbf{90.86} & \textbf{99.34}          & \textbf{95.33}    \\
            KLMatching   & 80.82          & 86.84          & 73.16          & 87.64          & 86.37          & 64.73          & 98.87          & 95.25    \\
            \bottomrule
        \end{tabular}
    \end{adjustbox}
    \caption{Results of OOD detection of two models of different families with and without the CIDER method applied. The models used are $Resnet_{18}$ and $ViT_{L16}$. The OOD detection results are reported for the SVHN dataset.}
    \label{table:results_ood_question3}
\end{table*}

\begin{table*}[!h]
    \centering
    \begin{adjustbox}{max width=\textwidth}
        \begin{tabular}{@{}lc*{19}cc@{}}
            \toprule
            \multirow{2}[3]{*}{\bfseries Dataset/Method} & 
            \multicolumn{2}{c}{\makecell{\bfseries SVHN }} & 
        \multicolumn{2}{c}{\makecell{\bfseries LSUNResize }} & 
        \multicolumn{2}{c}{\makecell{\bfseries Textures }} & 
        \multicolumn{2}{c}{\makecell{\bfseries TinyImageNetCrop }} & 
        \multicolumn{2}{c}{\makecell{\bfseries CIFAR100 }} \\
        \cmidrule(lr){2-3}\cmidrule(lr){4-5}\cmidrule(lr){6-7}\cmidrule(lr){8-9}\cmidrule(lr){10-11}\cmidrule(lr){12-13}\cmidrule(lr){14-15}
        & AUROC$\uparrow$ & ACC95TPR$\uparrow$ & AUROC$\uparrow$ & ACC95TPR$\uparrow$ & AUROC$\uparrow$ & ACC95TPR$\uparrow$ & AUROC$\uparrow$ & ACC95TPR$\uparrow$ & AUROC$\uparrow$ & ACC95TPR$\uparrow$\\
        \midrule
        Mahalanobis  & 6.72           & 83.59          & 20.86          & 48.00          & 48.38          & 35.62          & 16.85          & 47.66          & 26.57          & 79.26\\
        MaxLogit     & \textbf{88.79} & 91.64          & 92.25          & 83.18          & 86.44          & 67.91          & 91.79          & \textbf{83.60} & \textbf{87.00} & \textbf{88.70}\\
        MaxSoftmax   & 82.90          & 90.29          & 85.05          & 76.53          & 85.17          & 68.29          & 86.67          & 78.08          & 84.24          & 88.46\\
        ODIN         & 85.09          & 87.99          & 66.13          & 50.00          & 80.79          & 36.06          & 73.12          & 50.00          & 71.07          & 83.33\\
        OpenMax      & 82.62          & 87.76          & 90.15          & 79.16          & \textbf{88.47} & \textbf{71.34} & 91.77          & 82.27          & 83.25          & 86.82\\
        EnergyBased  & 88.73          & \textbf{91.71} & \textbf{92.37} & \textbf{83.47} & 86.43          & 68.32          & \textbf{91.87} & 83.54          & \textbf{87.00} & 88.68\\
        KLMatching   & 75.83          & 84.05          & 77.03          & 49.56          & 77.48          & 37.74          & 78.64          & 49.65          & 73.43          & 79.66\\
        \bottomrule
    \end{tabular}
    \end{adjustbox}
    \caption{Baseline results of OOD performance of the pre-trained convolutional model (resnet) without the CIDER method.}
    \label{table:results_resnet18_no_cider}
\end{table*}

\begin{table*}[!h]
    \centering
    \begin{adjustbox}{max width=\textwidth}
        \begin{tabular}{@{}lc*{17}cc@{}}
        \toprule
        \multirow{2}[3]{*}{\bfseries Detection Method} & 
        \multicolumn{2}{c}{\makecell{\bfseries Resnet18 }} & 
        \multicolumn{2}{c}{\makecell{\bfseries Resnet34 }} & 
        \multicolumn{2}{c}{\makecell{\bfseries Resnet50 }} & 
        \multicolumn{2}{c}{\makecell{\bfseries Resnet101 }} \\
        \cmidrule(lr){2-3}\cmidrule(lr){4-5}\cmidrule(lr){6-7}\cmidrule(lr){8-9}
        & AUROC$\uparrow$ & ACC95TPR$\uparrow$ & AUROC$\uparrow$ & ACC95TPR$\uparrow$ & AUROC$\uparrow$ & ACC95TPR$\uparrow$ & AUROC$\uparrow$ & ACC95TPR$\uparrow$ \\
        \midrule 
        Mahalanobis    & 5.38             & 83.59        &  5.45     & 83.59    &      5.48   & 83.59       & 4.94    &  83.59       \\
        MaxLogit       & \textbf{80.83}   & 90.19        &  78.14    & 89.66    &      77.83  & 89.81       & 76.92   &  89.45       \\
        MaxSoftmax     & 77.43            & 88.98        &  77.61    & 89.07    &      79.84  & 89.02       & 81.48   &  89.31       \\
        ODIN           & 80.01            & 87.99        &  85.10    & 87.99    &      83.96  & 87.99       & 87.21   &  90.01       \\
        OpenMax        & 85.66            & 88.06        &  \textbf{86.34}  & \textbf{90.25}  &       \textbf{90.62}& \textbf{91.25}     & \textbf{92.47} &  \textbf{91.89}     \\
        EnergyBased    & 80.75            & \textbf{90.23}      &  77.95    & 89.70    &      77.25  & 89.85       & 75.61   &  89.39       \\
        KLMatching     & 73.28            & 85.27               &  73.06    & 86.15    &      75.37  & 86.04       & 75.23   &  86.02       \\
        \bottomrule
    \end{tabular}
    \end{adjustbox}
    \caption{Out of domain detection results for different Resnet variants trained from scratch reported for the SVHN dataset}
    \label{table:resnet_scratch_size_variants}
\end{table*}

\begin{table*}[!h]
    \centering
    \begin{adjustbox}{max width=\textwidth}
        \begin{tabular}{@{}lc*{17}cc@{}}
        \toprule
        \multirow{2}[3]{*}{\bfseries Detection Method} & 
        \multicolumn{2}{c}{\makecell{\bfseries Resnet18 }} & 
        \multicolumn{2}{c}{\makecell{\bfseries Resnet34 }} & 
        \multicolumn{2}{c}{\makecell{\bfseries Resnet50 }} & 
        \multicolumn{2}{c}{\makecell{\bfseries Resnet101 }} \\
        \cmidrule(lr){2-3}\cmidrule(lr){4-5}\cmidrule(lr){6-7}\cmidrule(lr){8-9}
        & AUROC$\uparrow$ & ACC95TPR$\uparrow$ & AUROC$\uparrow$ & ACC95TPR$\uparrow$ & AUROC$\uparrow$ & ACC95TPR$\uparrow$ & AUROC$\uparrow$ & ACC95TPR$\uparrow$ \\
        \midrule 
        Mahalanobis    & 2.51  & 83.59 &     4.71  & 83.59 &    3.30 &  83.59         & 5.47  & 83.59     \\
        MaxLogit       & 82.44 & 91.08 &     76.01 & 89.56 &    73.44 & 89.45         & 78.56 & 89.56     \\
        MaxSoftmax     & 84.81 & 90.71 &     81.48 & 89.56 &    83.38 & 90.03         & 81.52 & 89.28     \\
        ODIN           & 83.19 & 87.99 &     78.83 & 87.99 &    83.40 & 87.99         & \textbf{89.40} & \textbf{90.40}     \\
        OpenMax        & \textbf{95.35} & \textbf{93.45} &     \textbf{93.26} & \textbf{91.10} &    \textbf{94.48} & \textbf{92.89}         & 84.61 & 90.27     \\
        EnergyBased    & 82.05 & 91.07 &     75.41 & 89.54 &    72.63 & 89.50         & 76.62 & 89.49     \\
        KLMatching     & 80.82 & 86.84 &     77.11 & 86.72 &    78.09 & 86.27         & 69.65 & 85.55     \\
        \bottomrule
    \end{tabular}
    \end{adjustbox}
    \caption{Out of domain detection results for different pre-trained Resnet variants reported for the SVHN dataset}
    \label{table:resnet_pretrained_size_variants}
\end{table*}

\begin{table*}[!h]
    \centering
    \begin{adjustbox}{max width=\textwidth}
        \begin{tabular}{@{}lc*{19}cc@{}}
        \toprule
        \multirow{2}[3]{*}{\bfseries Detection Method} & 
        \multicolumn{2}{c}{\makecell{$ViT_{B16}$ }} & 
        \multicolumn{2}{c}{\makecell{$ViT_{B32}$ }} & 
        \multicolumn{2}{c}{\makecell{$ViT_{L16}$ }} & 
        \multicolumn{2}{c}{\makecell{$ViT_{L32}$ }} \\
        \cmidrule(lr){2-3}\cmidrule(lr){4-5}\cmidrule(lr){6-7}\cmidrule(lr){8-9}
        & AUROC$\uparrow$ & ACC95TPR$\uparrow$ & AUROC$\uparrow$ & ACC95TPR$\uparrow$ & AUROC$\uparrow$ & ACC95TPR$\uparrow$ & AUROC$\uparrow$ & ACC95TPR$\uparrow$ \\
        \midrule    
        Mahalanobis    & 96.96 & 93.72 &     94.69 & 93.49 &    92.64 & 92.20   & \textbf{85.57} & \textbf{90.96} \\
        MaxLogit       & 99.46 & 95.39 &     98.22 & 94.78 &    97.20 & 94.24   & 85.36 & 88.89 \\
        MaxSoftmax     & 98.30 & 94.99 &     96.14 & 94.39 &    94.19 & 93.59   & 82.80 & 87.99 \\
        ODIN           & 71.66 & 87.15 &     67.97 & 86.48 &    90.85 & 87.99   & 71.99 & 87.99 \\
        OpenMax        & 95.57 & 94.83 &     94.10 & 94.35 &    95.35 & 94.05   & 85.19 & 89.22 \\
        EnergyBased    & \textbf{99.49} & \textbf{95.41} &     \textbf{98.30} & \textbf{94.79} &    \textbf{97.34} & \textbf{94.28}   & 85.39 & 88.89 \\
        KLMatching     & 98.10 & 94.97 &     95.82 & 94.36 &    84.73 & 91.37   & 75.50 & 90.17 \\
        \bottomrule
    \end{tabular}
    \end{adjustbox}
    \caption{Out of domain detection results for different Vision transformer variants reported for the SVHN dataset}
    \label{table:ood_vit_baseline}
\end{table*}

\subsection{Analysis}

In this section, we present our results with regards to the research questions asked in Section~\ref{sec:ood_method}. The different results are presented in the different tables below.

The baseline pipeline results in tables~\ref{table:results_ood_first_question} and~\ref{table:results_resnet18_no_cider} demonstrate the out-of-distribution performance of pre-trained ResNet and ViT models. These results attempt at answering the \hyperref[ood_research_question:1]{first research question}. We report the AUROC of the OOD classifiers and the Accuracy when the True Positive Ratio is 0.95.  As expected, the Mahalanobis performance of the ResNet was low due to convolutional models having a representation space that is not conducive to OOD detection. This result has sparked further research into improving OOD performance of CNN models. We however observe in our experiments that performances are improved on all the method tested when using the ViT model. As discussed in section~\ref{sec:ood_method}, the two points we are also interested in to confirm our experiments are the number  of parameters of each model and well as their in-domain performance.
This result leads us to the hypothesis that pre-trained transformer based-models are better suited for OOD detection due to either the nature of the model, pre-training pipeline or number of parameters. We also note that the best method depends on the OOD dataset used and some methods suffer a drastic loss in performance when tested on a specific dataset. For example, ODIN performs well on \textit{SVHN}: 86\% AUROC on average, but less on \textit{LSUNResize}: 66\% in both baselines.

To answer the \hyperref[ood_research_question:2]{second research question}, we compare the OOD detection performance of pre-trained CNN models with and without the CIDER finetuning. We are interested in knowing if the OOD detection performance is significantly influenced by the pre-trained nature of the model. To this end, we established a baseline with pre-trained resnet models. We chose different sizes of resnet models and report the OOD detection performance on the \textit{SVHN} dataset. These results, reported in table~\ref{table:resnet_scratch_size_variants} show us that the size of the Resnet model does not influence the OOD detection performance when training them from scratch on our ID dataset. Therefore, because it is the most computationally efficient, we use the $Resnet_{18}$ in the Table~\ref{table:results_ood_question2} to explore the use of the CIDER method on pre-trained CNN models on multiple OOD datasets. Although we only report the SVHN dataset, the rest of our experiments are accessible in the linked repository. We observe that, on several datasets, like \textit{SVHN}, the OOD detection performance increases when applying CIDER to the pre-trained model. However we notice a decrease of OOD detection performance on other metrics than Mahalanobis. This last metric, increases from 2.51 to 54.71 in the CIDER configuration. We can explain this behavior because of the nature of the CIDER method. Because CIDER modifies the nature of the latent space and the Mahalanobis metric is the only one to use the latent space directly and not the softmax output. In these circumstances, the CIDER method was beneficial and yielded an improved OOD detection performance than the baseline shown in Table~\ref{table:results_resnet18_no_cider}. We also observe that in some cases, on the \textit{Cifar100} OOD dataset for example, the OOD detection performance slightly drops compared to the baseline (from 87\% to 85.63\%). In our experiments, we managed to improve the OOD detection performance on some datasets and we notice that, on the dataset where the performances were reduced, the semantic similarities were the highest with the training dataset (\textit{Cifar10}).

Finally, to evaluate the \hyperref[ood_research_question:3]{third research question}, we compare the OOD performance results of pre-trained models form different families in different conditions. The results are reported in Table~\ref{table:results_ood_question3}. We observe that on datasets like SVHN as reported, the OOD detection performance is increased on all metrics when using a pre-trained $ViT_{L16}$ and further improved when using the CIDER method on it. When using the Mahalanobis metric, the OOD performance goes from 2.51 to a maximum of 97.48 in the \textit{ViT + CIDER} condition. From these results we conclude that using a vision transformer model is preferable to using a resnet for OOD detection. Further, we also tested this hypothesis with multiple variants of ViT models and reported the baseline results on the SVHN dataset in Table~\ref{ood:vit_variant_table}. 

Overall, we observe in table~\ref{table:ood_vit_baseline} and~\ref{table:results_ood_question3} that applying the CIDER method to a pre-trained ViT models yields good OOD results without sacrificing model performance in the same manner as in the original CIDER paper.
Furthermore, it was observed that pre-trained ViT models have a natural capacity for OOD detection without the need for fine-tuning. 
We obtain strong results when applying the CIDER method to pre-trained ViT models. The OOD detection performance is improved with all detection methods tested on almost all datasets. 

\subsection{Discussion}

We observed promising results for the research questions we asked in this paper. Mainly we saw that the CIDER method can be applied to pre-trained CNN and ViT and yield better OOD detection results on some datasets. However, we also saw the results were not as satisfactory in other cases. We explored several model variants for CNN and ViT to account for the role the number of parameters may have in our results. In addition, we also report the in-domain performance of each network on our repository as it may have a detrimental role on OOD detection performances. 

Although the method we used yields positive results, it is too early to conclude on its efficiency. The results presented in this paper should be further confirmed for other in-domain datasets of increasing complexity. Neural network do not necessarily scale to more complex datasets and tasks, hence the need for further investigations with other Computer Vision tasks such as image segmentation.

    


\section{Conclusion} 


We empirically explored OOD performances of pre-trained CNN and ViT models. In our experiments we tested different models of the resnet and ViT families in different conditions on different out of domain datasets with different detection methods. 


Our empirical exploration of OOD performances of pre-trained convolutional neural network and vision transformer models reveals that ViT models have superior out-of-the-box performances than their CNN counterparts. We showed that the CIDER method generalizes to pre-trained Resnet model and improve their OOD performance only when using the Mahalanobis detection method. Furthermore, applying CIDER to a pre-trained ViT improved OOD detection scores on all detection methods consistently. 

Initial results suggest that the ViT model has an interesting capability to detect out of domain data, however, the dataset used in this experiment is not complex enough to make a general conclusion about this property. Thus, further testing on larger datasets is necessary to determine the source of this behavior. Additionally, future works should explore the architecture difference and pre-training pipeline of the models to gain further insight into their OOD detection capabilities.


\printbibliography 

\end{document}